\documentclass{article}
\usepackage{spconf,amsmath,graphicx}
\newcommand{\comment}[1]{}

\usepackage{multirow}
\usepackage{subcaption}

\newcommand{\bxi}{\boldsymbol{\xi}}

\usepackage{booktabs}
\usepackage{makecell} 
\usepackage{color}
\usepackage{xcolor}
\usepackage{amssymb}
\usepackage{enumitem}
\setlist{nosep, leftmargin=14pt}

\usepackage{mwe} 

\title{NeRFscopy: Neural Radiance Fields for in-vivo time-varying\\ tissues from endoscopy}

\name{Laura Salort-Benejam and Antonio Agudo\thanks{This work has been supported by the project GRAVATAR PID2023-151184OB-I00 funded by MCIU/AEI/10.13039/501100011033 and by ERDF, UE; and by the Government of Catalonia under 2025 FI-STEP 00398.}}
\address{Institut de Rob\`otica i Inform\`atica Industrial, CSIC-UPC, Spain}

\begin{document}
%
\maketitle
\begin{abstract}
Endoscopy is essential in medical imaging, used for diagnosis, prognosis and treatment. Developing a robust dynamic 3D reconstruction pipeline for endoscopic videos could enhance visualization, improve diagnostic accuracy, aid in treatment planning, and guide surgery procedures. However, challenges arise due to the deformable nature of the tissues, the use of monocular cameras, illumination changes, occlusions and unknown camera trajectories. Inspired by neural rendering, we introduce NeRFscopy, a self-supervised pipeline for novel view synthesis and 3D reconstruction of deformable endoscopic tissues from a monocular video. NeRFscopy includes a deformable model with a canonical radiance field and a time-dependent deformation field parameterized by SE(3) transformations. In addition, the color images are efficiently exploited by introducing sophisticated terms to learn a 3D implicit model without assuming any template or pre-trained model, solely from data. NeRFscopy achieves accurate results in terms of novel view synthesis, outperforming competing methods across various challenging endoscopy scenes.
\end{abstract}
\begin{keywords}
Neural rendering, Non-rigid tissues, Endoscopy, 3D scene understanding, Monocular Vision.
\end{keywords}

\section{Introduction}

Endoscopy is a widely used imaging method for examining internal organs and cavities for diagnostic, prognostic, or therapeutic purposes. Maybe, the most standard way is to use a monocular endoscope due to its compact size and versatility~\cite{stereo_not_common}. Since endoscopes are used in very different interventions, such as gastroscopies, laparoscopies or bronchoscopies, it is mandatory to consider models that are capable of solving the problem for this great diversity of visual information and tissue deformations. As a consequence, our goal is to propose \textit{a generic and universal method for endoscopy understanding} that may benefit both patients and clinicians. For example, it could facilitate the assessment of various potentially harmful structures, accurately provide information on the shape and dimensions of nodules or other anatomical clues, allow physicians to visualize the structures in 3D and also obtain novel views of those same structures after the procedure is completed. Thus, allowing them to make more informed decisions on diagnosis or treatment, or for future check-ups in which the acquisition of measurements could be compared to a newer one to determine the progress of a disease.

In the last decade, recovering 3D information from monocular images has been extensively studied. Perhaps the most important family of algorithms to address this problem is structure from motion (SfM) in its variants for rigid~\cite{schoenberger2016sfm} and non-rigid objects~\cite{AgudoPAMI2016,agudoICPR2020}, and others such as shape from template~\cite{CasillasIJCV2021}, shape from shading~\cite{LightDepth}, photometric stereo~\cite{photometric-stereo} or supervised learning~\cite{bimodal_camera_pose}. Previous approaches exploit the use of explicit correspondences between images to solve the problem in combination with the use of priors that limit the type of motion and deformation of tissues. Dealing with deformable objects, as is the case in most in vivo tissues, is more challenging because the problem is inherently ill-posed. Endoscopy videos present additional challenges that can produce outliers and decrease the robustness of shape and camera estimations, such as specularities on the tissue, sparse viewpoints, occlusions by the endoscope or bodily fluids, lack of texture, unexpected camera movements, and motion blur. The recent development of Neural Radiance Fields (NeRF)~\cite{original-NERF}, a rendering approach that implicitly represents 3D volumes with a neural network, has shown promise in overcoming the limitations of traditional 3D reconstruction methods in different scenarios, but was
originally intended for rigid objects and under acceptable visual conditions. Another development is 3D Gaussian Splatting~\cite{gaussiansplatting}, which builds upon the concept of radiance fields and compactly represents the scene with a set of 3D Gaussians. However, it is limited for rigid scenes and requires SfM for camera calibration and initialization of the scene's geometry. Later, some works have adapted these two recent methods for endoscopic videos~\cite{endogaussian,endoNERF,lerplane,endoSURF}.

In this paper, we propose a self-supervised NeRF-based approach for novel view synthesis of time-varying biological tissues from a generic monocular endoscopy video. This information is exploited, without relying on a priori knowledge, to learn a volumetric representation of the non-rigid tissues by using sophisticated terms that ensure accurate results. We show superiority over previous works focused on endoscopy.

\section{NeRF for dynamic endoscopy tissues}

\begin{figure}[t!]
    \centering
    \includegraphics[width=0.99\linewidth]{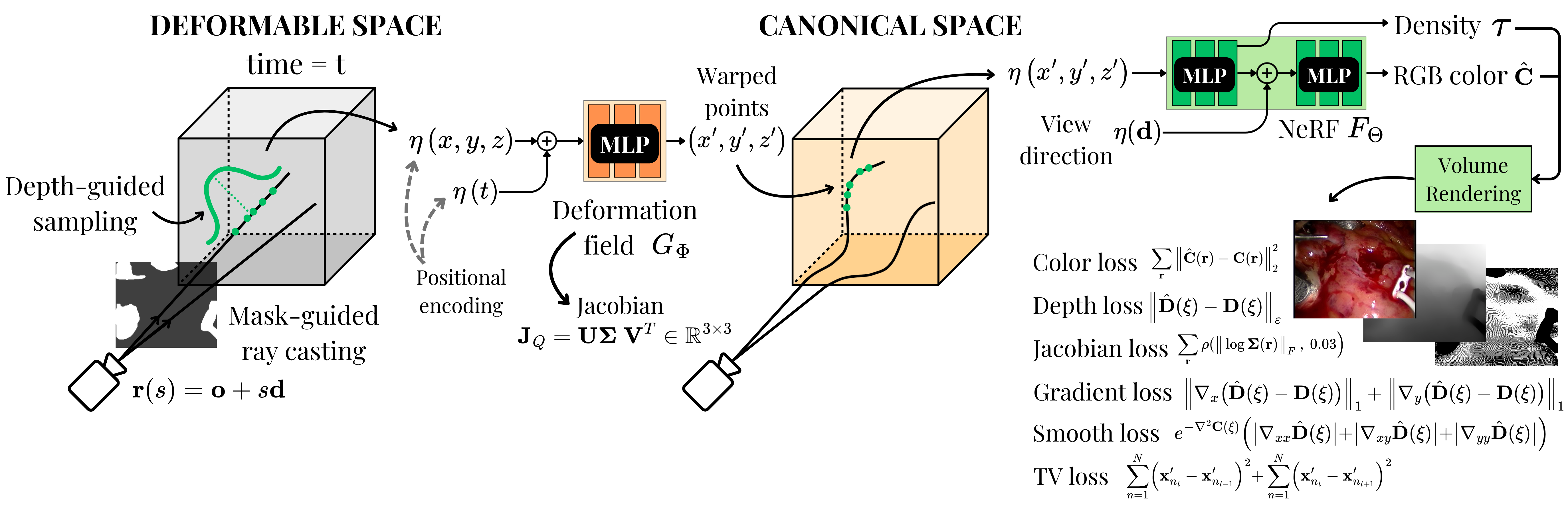}
    \vspace{-0.4cm}
    \caption{\small{\textbf{Overview of our NeRFscopy} pipeline for time-varying NeRF from RGB endoscopy videos.}}
    \label{fig:model-overview}
\end{figure}

The fundamental concept behind any NeRF~\cite{original-NERF} method is the implicit continuous representation of a scene through the weights $\Theta$ of a multilayer perceptron (MLP) $F_\Theta$. This network takes a 5D vector as input, containing 3D coordinates $\textbf{x}=(x,y,z)$, obtained by marching camera rays through the scene and 2D viewing direction $(\theta,\phi)$ of the camera, which is expressed as a 3D Cartesian unit vector $\textbf{d}$. The input vector is then mapped to an emitted radiance $\textbf{c}=(r,g,b)$ and a volume density $\tau$ as $F_\Theta : (\textbf{x},\textbf{d}) \xrightarrow{} (\textbf{c},\tau)$. With these ingredients, the previous function can represent the shape as the volume density and directional emitted radiance at any 3D point. A point along the ray $\textbf{r}(s)=\textbf{o} +s\textbf{d}$ is defined by its origin $\textbf{o}$ and direction $\textbf{d}$, being $s$ a scalar to encode a particular location in the ray. In this work, we propose NeRFscopy, a neural model that introduces an SE(3) encoding of the deformations together with sophisticated terms to improve the implicit representation of a scene (see our pipeline in Fig.~\ref{fig:model-overview}). To this end, our model exploits an RGB video $(\textbf{I}_i)_{i=1}^I \in \mathbb{R}^{H \times W \times 3}$ as input that is acquired by a calibrated camera, where $H$, $W$, and $I$ represent height, width, and the total number of frames, respectively. Thanks to that, our method is able to synthesize novel views with high accuracy, while implicitly knowing the 3D shape of the observed object. To avoid prior knowledge of rigid points in the scene with which to disambiguate between rigid scene and camera motion, we will assume that the camera motion is null, since our interest is to capture the largest possible number of non-rigid vivid tissues without considering a segmentation of motions in advance. In any case, our method could easily exploit that motion when available. Without loss of generality, and following non-rigid approaches in NeRF~\cite{4DPV}, we just consider the object of interest by defining a binary mask $\textbf{M}_i \in \mathbb{R}^{H \times W}$ that excludes the surgical tools.

Inspired by EndoNeRF~\cite{endoNERF}, we exploit a depth-guided sampling with a Gaussian transfer function near the estimated surface of the tissue. In order to adapt this work for monocular inputs, we propose the use of pre-trained monocular depth estimation algorithms, obtaining for the $i$-th image a relative depth map $\textbf{D}_i \in \mathbb{R}^{H \times W}$. For the deformable endoscopy scene representation, and following~\cite{nerfies,4DPV}, we propose to use a canonical radiance field $F_\Theta$ together with a time-dependent deformation one $G_\Phi$. However, instead of using a displacement field, as in~\cite{endoNERF}, we draw inspiration from~\cite{nerfies} and propose to employ a dense SE(3) deformation field that encodes a rigid transformation. This decision is based on the notion that a simple displacement field computes a different translation for each point. While being able to represent smooth deformations, it is not sufficient to capture complex and simultaneous rotations of points across different regions of the scene, which can be efficiently achieved by the SE(3) deformation field using significantly less parameters.

\textbf{Deformable network.} The deformation field consists of an 8-layer MLP $G_{\Phi}:(\textbf{x},t)\rightarrow (\textbf{a}, \textbf{b})$ that encodes a rigid transformation between the input point $\textbf{x}$ at time \textit{t} of the \textit{i}-th frame normalized as $t_i = i/I$, and the corresponding point in the canonical space through a screw axis $ \mathcal{S} = (\frac{\textbf{a}}{\| \textbf{a} \|} , \frac{\textbf{b}}{\| \textbf{b} \|})=(\hat{\textbf{a}},\hat{\textbf{b}})\in \mathbb{R}^6$, being $\hat{\textbf{a}}$ the axis of rotation in $\mathfrak{so}(3)$ --the Lie algebra of SO(3)-- and $\| \textbf{a}\|$ the angle of rotation $\zeta$. The full rotation matrix $\textbf{R} \in SO(3)$ is obtained by means of the Rodrigues' formula as $\textbf{R}= \textbf{I}_3+[\textbf{A}]\sin{\zeta} + [\textbf{A}]^2 (1-\cos{\zeta})$, where $\textbf{A}$ is a skew-symmetric matrix of $\hat{\textbf{a}}$ and $\textbf{I}_3$ a 3$\times$3 identity matrix. Following~\cite{screwaxis}, the translation component $\textbf{p}\in \mathbb{R}^3$ can be calculated as $\textbf{p}=(\zeta\textbf{I}_3 + [\textbf{A}](1-\cos{\zeta}) + \bigr[\textbf{A}]^2(\zeta - \sin{\zeta}))[\hat{\textbf{b}}]$. The warped points in the canonical space are computed as $ \textbf{x}'=\textbf{Q}\textbf{x}= \textbf{R}\textbf{x}+\textbf{p}$, where $\textbf{Q}=\begin{bmatrix}
        \textbf{R}&\textbf{p}\\
        \textbf{0}& 1
    \end{bmatrix} \in SE(3) $
encodes a homogeneous rigid transformation. The color and volume density are then obtained as in NeRF~\cite{original-NERF} with the exception that NeRFscopy is composed only of one MLP instead of using coarse and fine models. This is due to the fact that using depth-guided sampling ensures that the samples are concentrated around what is estimated to be the surface of the tissues, removing so the need for hierarchical volume sampling.

\textbf{Optimization.} NeRFscopy learns by minimizing a differentiable function that is optimized using stochastic gradient descent in a self-supervised manner. We next define the loss function that we propose to solve the problem. Following differentiable rendering approaches~\cite{original-NERF}, we use a photometric loss and also a depth loss to penalize the deviation between the predicted depth map $\hat{\textbf{D}}(\textbf{r})$ and the pre-computed one as: 
\begin{equation}
\mathcal{L}_{C}= \sum_{\textbf{r}} \| \hat{\textbf{C}}(\textbf{r}) -\textbf{C}(\textbf{r}) \|_2^2, \quad \mathcal{L}_{D}= \| \hat{\textbf{D}}(\bxi) -\textbf{D}(\bxi) \|_\varepsilon,
\label{photo_loss}
\end{equation}
where $\hat{\textbf{C}}(\textbf{r})$ and $\textbf{C}(\textbf{r})$ are rendered and observed pixel colors, respectively, $\bxi$ includes all the point rays and $\varepsilon$ denotes a Huber norm. 
To enforce a local deformation of $G_\Phi$ we propose to penalize the deviation from zero of the logarithm of the singular values $\boldsymbol{\Sigma}$ of the Jacobian matrix (see Fig.~\ref{fig:model-overview}) obtained through automatic differentiation of $G_\Phi$~\cite{nerfies} as:
\begin{equation}
\mathcal{L}_{\textbf{}J}=\sum_{\textbf{r}} \rho(\|\log\boldsymbol{\Sigma}(\textbf{r}) \|_F \ , \ c),
\label{jac_loss}
\end{equation}
where $\rho(\cdot)$ is a Geman-McClure robust error function~\cite{mcclure}, 
to reduce the impact of non-rigid motions that would hinder the training of the deformation field. $\|\cdot\|_F$ is a Frobenius norm. Inspired by~\cite{gradient-depth}, we propose a depth gradient regularization to encourage the discontinuities in the estimated depth to be sharp and similar to those in the input depth, defined as the $l_1$ norm between the gradients of the difference between the estimated and input depths as $\mathcal{L}_{g}=\| \nabla_x (\hat{\textbf{D}}(\bxi) - \textbf{D}(\bxi)) \|_1 + \| \nabla_y (\hat{\textbf{D}}(\bxi) - \textbf{D}(\bxi)) \|_1$ .

We also incorporate a depth smoothness loss~\cite{smooth-loss}, enforcing neighboring pixels to have similar depth values by using second-order gradients of the estimated depth as:
\begin{equation}
\mathcal{L}_{s}= e^{-\nabla^2\textbf{C}(\bxi)} (| \nabla_{xx} \hat{\textbf{D}}(\bxi)| + |\nabla_{xy} \hat{\textbf{D}}(\bxi) |+ |\nabla_{yy}\hat{\textbf{D}}(\bxi)| ),
\end{equation}
where $\nabla^2\textbf{C}(\bxi)$ is the Laplacian of the sampled input color, whose exponential is used as a weighting term to assign less importance to pixels that are more likely to be edges and discontinuities. Finally, we add a temporal total variation (tv) regularization to enforce the deformation field of consecutive time instances to be similar, resulting in temporally coherent displacements and avoiding abrupt changes:
\begin{equation}
\mathcal{L}_{tv}=\sum_{n=1}^N ( \textbf{x}'_{n_t} - \textbf{x}'_{n_{t-1}})^2 + \sum_{n=1}^N ( \textbf{x}'_{n_t} - \textbf{x}'_{n_{t +1}})^2 . 
\end{equation}
This term is defined as the $l_2$ norm between a set of warped points $\textbf{x}'_{n_t}$ in the canonical space at the current time $t$ and the warped ones $\textbf{x}'_{n_{t-1}}$ and $\textbf{x}'_{n_{t+1}}$, at a previous and following time instances $t-1$ and $t+1$, respectively. Color and depth representations are optimized by minimizing the total loss:
\begin{equation}
    \mathcal{L} = \mathcal{L}_{C} +\lambda_1 \mathcal{L}_{D} + \lambda_2 \mathcal{L}_{J} + \lambda_3\mathcal{L}_{g}    
    + \lambda_4 \mathcal{L}_{s} + \lambda_5 \mathcal{L}_{tv},
    \label{final-loss}
\end{equation}
where $\lambda_1$, $\lambda_2$, $\lambda_3$, $\lambda_4$ and $\lambda_5$ are weighting factors.

\section{Experimental Evaluation}

We now present our experimental results on four in-vivo monocular videos featuring mild to severe deformations, diverse textures, and illumination-varying conditions, where 70 consecutive frames per video were selected. These include two Totally Endoscopic Coronary Artery Bypass (TECAB) surgeries~\cite{AgudoICIP2021,heart} with resolution 360$\times$288 and 348$\times$576, respectively; a lung lobectomy~\cite{hamlynsurgeryvid}, with resolution 348$\times$576; and a bronchoscopy~\cite{UrdapilletaICIP2023}, with resolution 320$\times$256. Additionally, we also consider the Endo-NeRF~\cite{endoNERF} dataset which is composed of two 640$\times$512 robotic prostatectomy videos.

For quantitative evaluation, we provide a Peak Signal-to-Noise Ratio (PSNR)~\cite{psnr}, Structural Similarity Index Measure (SSIM)~\cite{ssim} and Learned Perceptual Image Patch Similarity (LPIPS)~\cite{lpips}, including a comparison, when possible, with EndoNeRF~\cite{endoNERF}, EndoSurf~\cite{endoSURF}, LerPlane-32k~\cite{lerplane} and EndoGaussian-monocular~\cite{endogaussian}. To make a fair comparison, in our implementation of NeRFscopy we adopted the same parameters used in EndoNeRF~\cite{endoNERF} in terms of the number of rays, samples per ray and the depth loss weight $\lambda_1$. In addition, the rest of the weights in Eq.~\eqref{final-loss} are set as $\lambda_2 =1e-6$, $\lambda_3=1.0$, $\lambda_4=1e-2$ and $\lambda_5=1e-4$, using the same values in all the experiments.

\newcolumntype{?}[1]{!{\vrule width #1}}
\begin{table}[t!]
    \centering
    \resizebox{7.5cm}{!} {
    \begin{tabular}{l|l|c|c|c}
         \multicolumn{2}{c|}{}& {PSNR} $\uparrow$ & {SSIM} $\uparrow$& {LPIPS}$\downarrow$\\\hline
         \multirow{3}{*}{TECAB1} & DPT~\cite{DPT}& \textbf{26.096}& \textbf{0.767}& 0,248\\\cline{2-5}
         &IID-SfmLearner~\cite{IID_SfmLearner}& 26.054& 0.761& \textbf{0.246}\\\cline{2-5}
         &Depth-Anything~\cite{depthanything}& 25.743& 0.750& 0.263\\
        \hline
         \hline
         \multirow{3}{*}{TECAB2} &DPT~\cite{DPT}& 24.716& 0.676&0.426\\\cline{2-5}
         &IID-SfmLearner~\cite{IID_SfmLearner}& \textbf{24.938}& \textbf{ 0.692}&\textbf{0.418}\\\cline{2-5}
         &Depth-Anything~\cite{depthanything}& 24.904& 0.678&0.423 \\ 
         \hline
         \hline
         \multirow{3}{*}{Lung Lobectomy}&DPT~\cite{DPT}& \textbf{27.782}& \textbf{ 0.796}&0.275\\\cline{2-5}
         &IID-SfmLearner~\cite{IID_SfmLearner}& 27.574& 0.796&0.277\\\cline{2-5}
         &Depth-Anything~\cite{depthanything}& 27.612& 0.795&\textbf{0.274}\\
         \hline
         \hline
         \multirow{3}{*}{Bronchoscopy}&DPT~\cite{DPT}& 33.589& \textbf{0.862}&\textbf{0.173}\\\cline{2-5}
         &IID-SfmLearner~\cite{IID_SfmLearner}& \textbf{33.713}& 0.861&0.186\\\cline{2-5}
         &Depth-Anything~\cite{depthanything}& 33.499& 0.861&0.186\\
         \bottomrule
            \end{tabular}}
            \vspace{-0.2cm}
    \caption{\small{\textbf{Depth term evaluation} on four real videos by using 3 monocular depth estimators (DPT~\cite{DPT}, IID-SfmLearner~\cite{IID_SfmLearner} and Depth-Anything~\cite{depthanything}) in our NeRFscopy.}\label{tab:ablation-depth}
    \label{tab:ablation-depth}}
    \end{table}

\begin{figure}[t!]
\centering
\begin{tabular}{@{}cccc@{}}
    \includegraphics[clip, angle=0, height=0.19\linewidth]{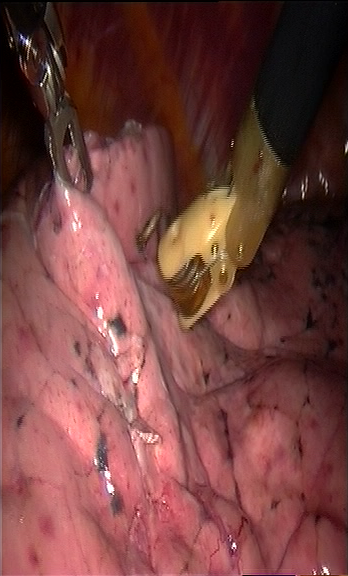}&
    \hspace{-0.45cm}
    \includegraphics[clip, angle=0, height=0.19\linewidth]{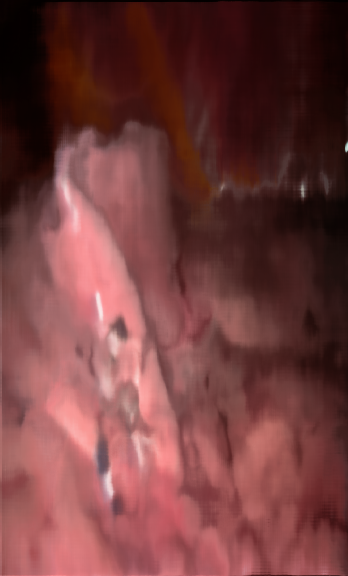}&
    \hspace{-0.45cm}
    \includegraphics[clip, angle=0, height=0.19\linewidth]{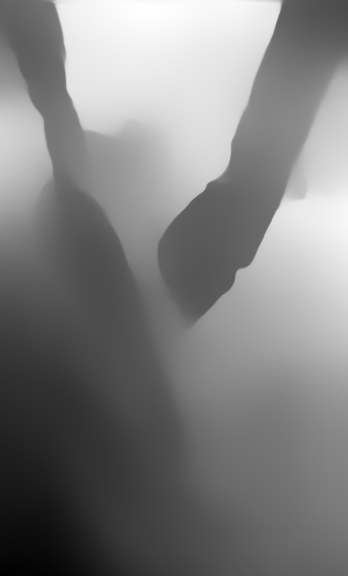}&
    \hspace{-0.45cm}
    \includegraphics[clip, angle=0, height=0.19\linewidth]{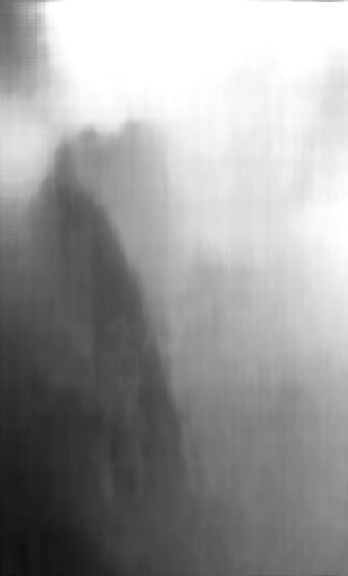}\vspace{-0.07cm}\\
    
 \includegraphics[clip, angle=0, height=0.19\linewidth]{Figures/ablation/surgery3/80.png}&
  \hspace{-0.45cm}
  \includegraphics[clip, angle=0, height=0.19\linewidth]{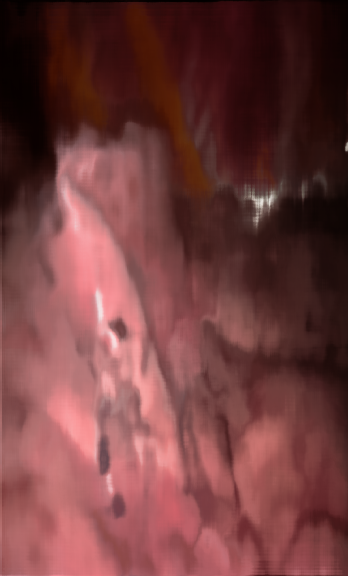}&
   \hspace{-0.45cm}
 \includegraphics[clip, angle=0, height=0.19\linewidth]{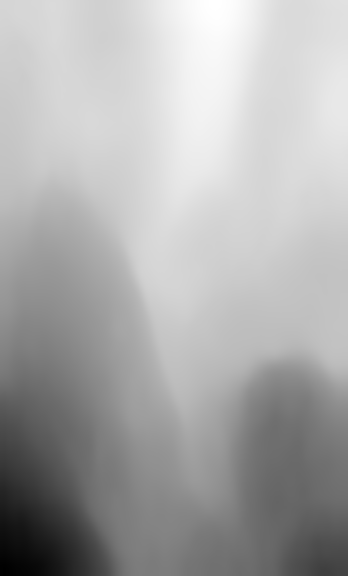}&
  \hspace{-0.45cm}
 \includegraphics[clip, angle=0, height=0.19\linewidth]{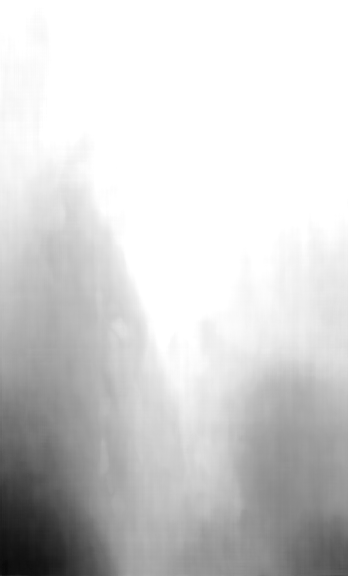}\vspace{-0.07cm}\\

  \includegraphics[clip, angle=0, height=0.19\linewidth]{Figures/ablation/surgery3/80.png}&
  \hspace{-0.45cm}
  \includegraphics[clip, angle=0, height=0.19\linewidth]{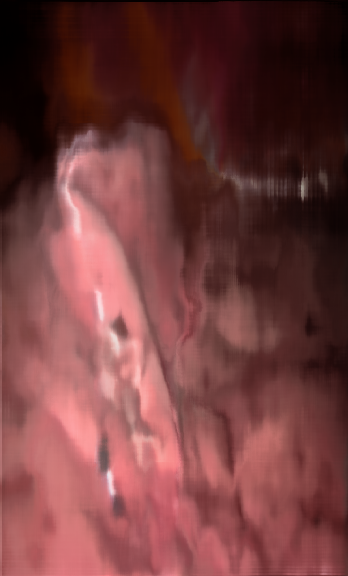}&
   \hspace{-0.45cm}
 \includegraphics[clip, angle=0, height=0.19\linewidth]{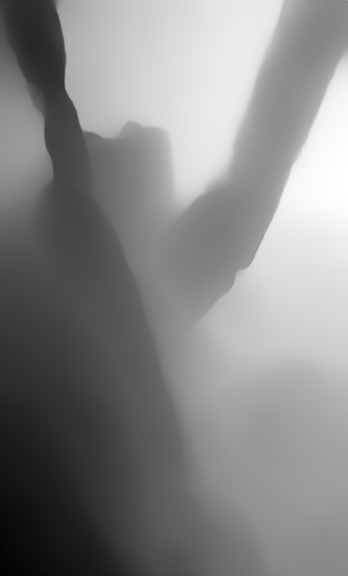}&
  \hspace{-0.45cm}
 \includegraphics[clip, angle=0, height=0.19\linewidth]{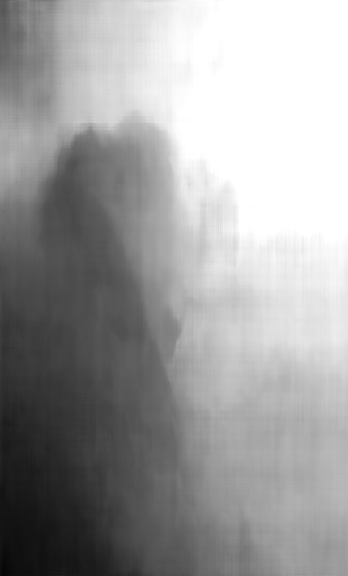}\vspace{-0.07cm}\\
\end{tabular}
\vspace{-0.3cm}
   \caption{\small{\textbf{Qualitative evaluation of the depth regularizer on the lung lobectomy video.} Columns show from left to right: arbitrary input frame, RGB rendered image, input depth estimation and depth rendered result. For depth initialization, we consider DPT~\cite{DPT} (top), IID-SfmLearner~\cite{IID_SfmLearner} (middle) or Depth-Anything~\cite{depthanything} (bottom).}}
\label{fig:depth-surgery3}
\end{figure}

\textbf{Depth term analysis.} We first propose to analyze the effect of the depth regularizer and its ability to include details in the final estimation of NeRFscopy. To do that, in this work we consider three pre-trained models, including: DPT~\cite{DPT}, IID-SfmLearner~\cite{IID_SfmLearner}, and Depth-Anything~\cite{depthanything}; which cover global models as well as more specialized ones in medical videos. Our results are reported in Table~\ref{tab:ablation-depth}. On average, the best results are obtained with DPT~\cite{DPT}, even though the numerical performance is quite similar. We then propose a visual assessment (see some examples in Fig.~\ref{fig:depth-surgery3}), and observing that Depth-Anything~\cite{depthanything} provided more visually accurate and detailed results we decided to use it to model $\textbf{D}(\textbf{r})$.

\begin{table}[t!]
    \centering
    \resizebox{8cm}{!} {
    \begin{tabular}{l|l|c|c|c}
         \multicolumn{2}{c|}{}& {PSNR} $\uparrow$ &  {SSIM} $\uparrow$& {LPIPS}$\downarrow$\\\hline
         \multirow{5}{*}{TECAB1} & EndoNeRF~\cite{endoNERF}& 25.791&  0.742& 0.255\\\cline{2-5}
         &NeRFscopy baseline&  25.743&  0.750& 0.263\\\cline{2-5}
         &NeRFscopy w/ grad&\textbf{25.874}&  \textbf{0.757}& \textbf{0.250}\\\cline{2-5}
         &NeRFscopy w/ grad, smooth& \underline{25.811}& \underline{0.750}& \underline{0.255}\\ \cline{2-5} 
         &NeRFscopy w/ grad, smooth, tv&  25.137&  0.743& 0.274\\
         \hline
         \hline
         \multirow{5}{*}{TECAB2} &EndoNeRF~\cite{endoNERF}& \textbf{24.954}&  0.685& \textbf{0.419}\\\cline{2-5}
         &NeRFscopy baseline&  \underline{24.904}&  0.678& 0.423\\\cline{2-5}
         &NeRFscopy w/ grad&  24.875& \underline{0.685}& \underline{0.423}\\\cline{2-5}
         &NeRFscopy w/ grad, smooth&  24.864&\textbf{0.689}& 0.429\\\cline{2-5}
         &NeRFscopy w/ grad, smooth, tv&  24.159&  0.676& 0.431\\ 
         \hline
         \hline
         \multirow{5}{*}{Lung Lobectomy} &EndoNeRF~\cite{endoNERF}& 27.142&  0.788& 0.293\\\cline{2-5}
         &NeRFscopy baseline&  \textbf{27.612}&	 \textbf{0.795}& \textbf{0.274}\\\cline{2-5}
         &NeRFscopy w/ grad& \underline{27.475}& \underline{0.793}& 0.277\\\cline{2-5}
         &NeRFscopy w/ grad, smooth&  27.285&  0.791& \underline{0.275}\\\cline{2-5}
         &NeRFscopy w/ grad, smooth, tv&  27.284&  0.792& 0.282\\
            \hline
         \hline
         \multirow{5}{*}{Bronchoscopy} &EndoNeRF~\cite{endoNERF}& 33.872&  0.867& 0.588\\\cline{2-5}
         &NeRFscopy baseline&  \underline{34.076}& \underline{0.873}& \underline{0.160}\\\cline{2-5}
         &NeRFscopy w/ grad&  33.499&  0.861& 0.186\\\cline{2-5}
         &NeRFscopy w/ grad, smooth&  \textbf{34.405}&  \textbf{0.875}&\textbf{ 0.156}\\\cline{2-5}
         &NeRFscopy w/ grad, smooth, tv&  32.579&  0.850& 0.201\\
         \bottomrule
            \end{tabular}}
            \vspace{-0.2cm}
            \caption{\small{\textbf{Ablation study of NeRFscopy and comparison with EndoNeRF~\cite{endoNERF}.} Each prior in Eq.~\eqref{final-loss} is evaluated with the NeRFscopy baseline. Results shown as {\textbf{best}} and \underline{second best}.}}
            \label{tab:ablation-losses}
            \vspace{-0.3cm}
        \end{table}

\begin{figure}[t!]
\centering
\resizebox{7.0cm}{!} { 
\begin{tabular}{@{}ccccc@{}}
    \includegraphics[clip, angle=0, height=0.17\linewidth]{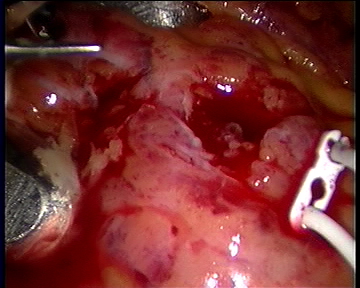}&
    \hspace{-0.45cm}
    \includegraphics[clip, angle=0, height=0.17\linewidth]{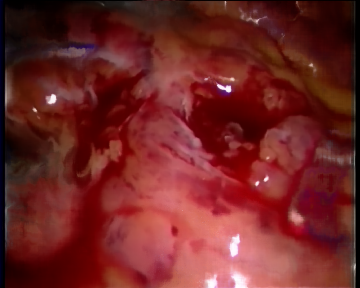}&
    \hspace{-0.45cm}
    \includegraphics[clip, angle=0, height=0.17\linewidth]{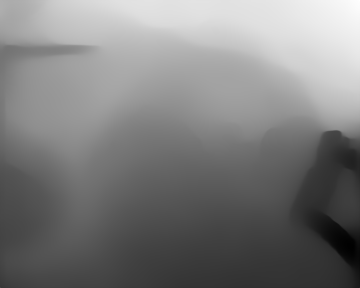}&
    \hspace{-0.45cm}
    \includegraphics[clip, angle=0, height=0.17\linewidth]{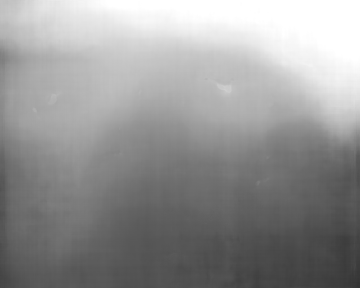}&
    \hspace{-0.45cm}
    \includegraphics[clip, angle=0, height=0.17\linewidth]{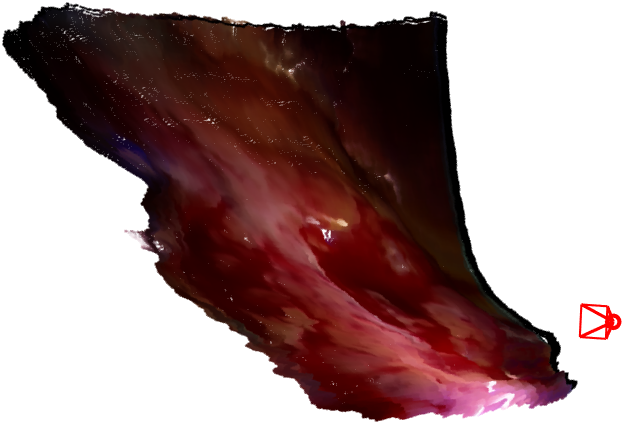}\vspace{-0.07cm}\\
  \includegraphics[clip, angle=0, height=0.35\linewidth]{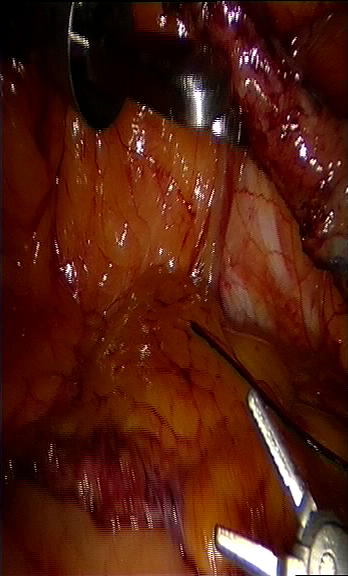}&
 \hspace{-0.45cm}
  \includegraphics[clip, angle=0, height=0.35\linewidth]{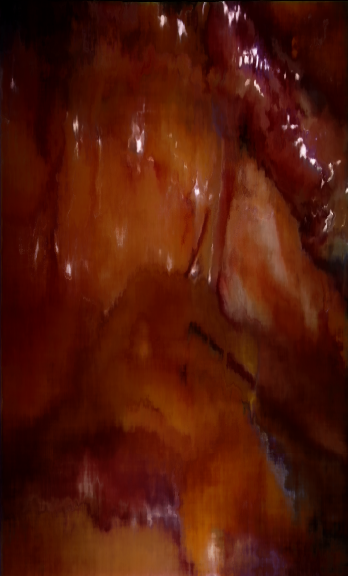}&
  \hspace{-0.45cm}
 \includegraphics[clip, angle=0, height=0.35\linewidth]{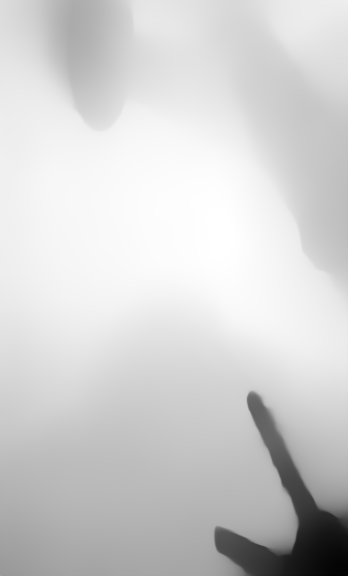}&
 \hspace{-0.45cm}
 \includegraphics[clip, angle=0, height=0.35\linewidth]{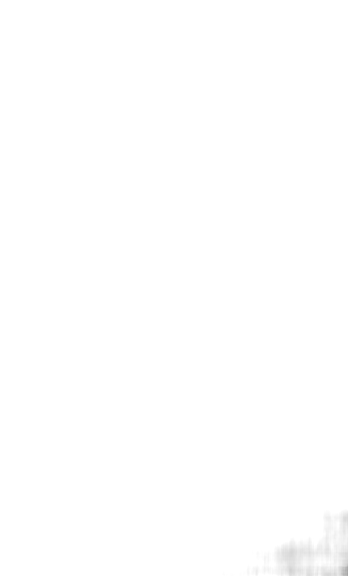}&
    \hspace{-0.45cm}
    \includegraphics[clip, angle=0, height=0.35\linewidth]{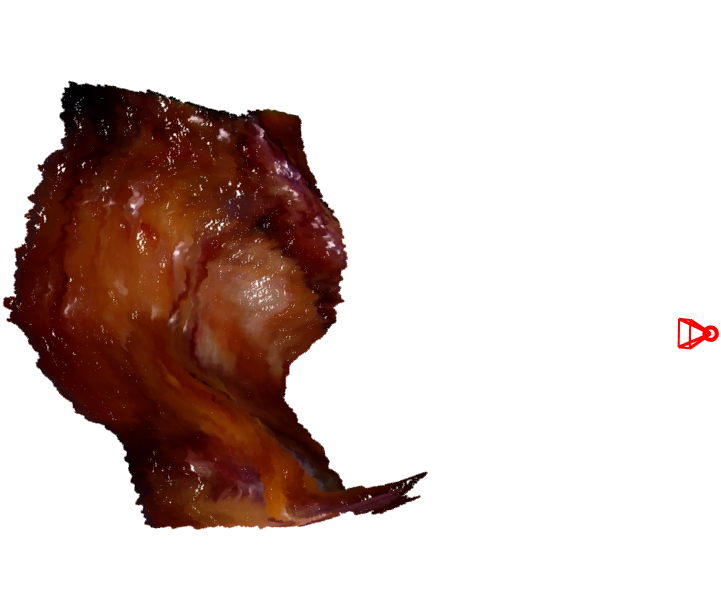}\vspace{-0.07cm}\\
 \includegraphics[clip, angle=0, height=0.35\linewidth]{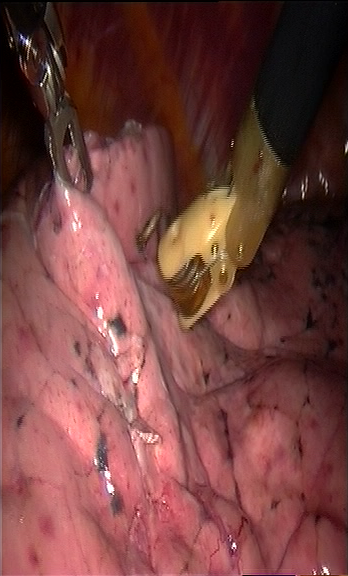}&
 \hspace{-0.45cm}
  \includegraphics[clip, angle=0, height=0.35\linewidth]{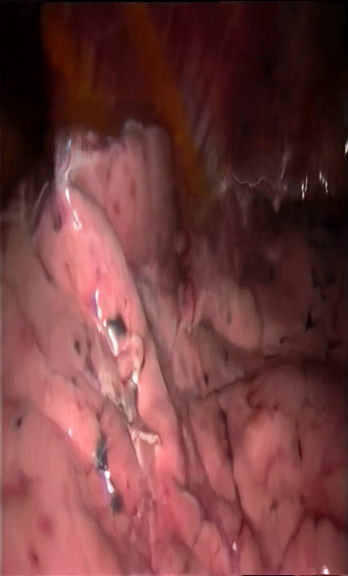}&
  \hspace{-0.45cm}
 \includegraphics[clip, angle=0, height=0.35\linewidth]{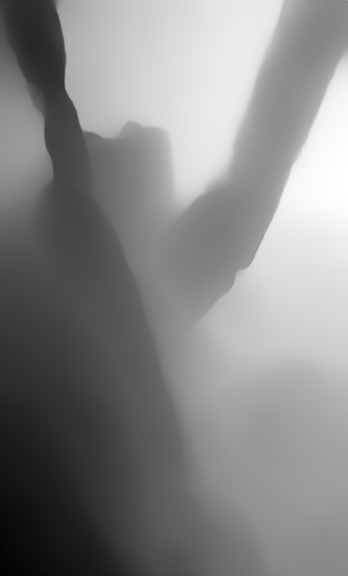}&
 \hspace{-0.45cm}
 \includegraphics[clip, angle=0, height=0.35\linewidth]{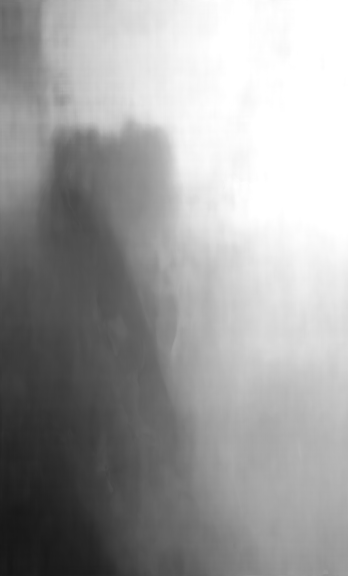}&
    \hspace{-0.45cm}
    \includegraphics[clip, angle=0, height=0.35\linewidth]{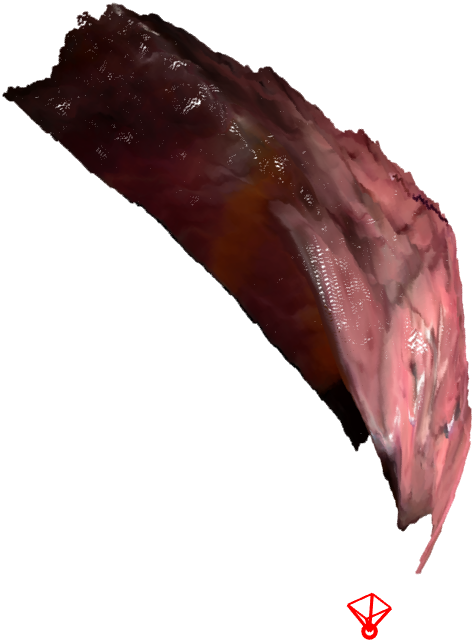}\vspace{-0.07cm}\\

  \includegraphics[clip, angle=0, height=0.17\linewidth]{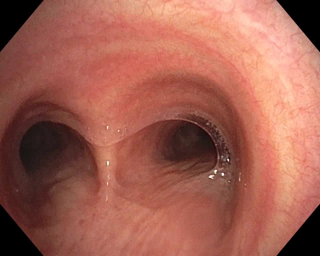}&
 \hspace{-0.45cm}
  \includegraphics[clip, angle=0, height=0.17\linewidth]{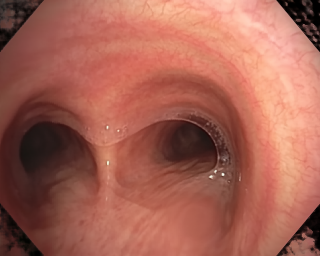}&
  \hspace{-0.45cm}
 \includegraphics[clip, angle=0, height=0.17\linewidth]{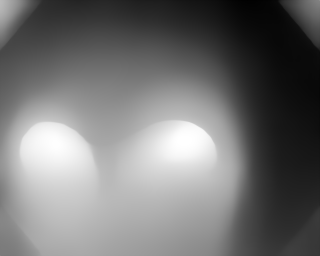}&
 \hspace{-0.45cm}
 \includegraphics[clip, angle=0, height=0.17\linewidth]{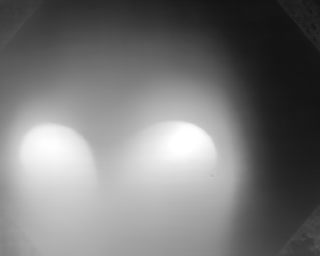}&
    \hspace{-0.45cm}
    \includegraphics[clip, angle=0, height=0.17\linewidth]{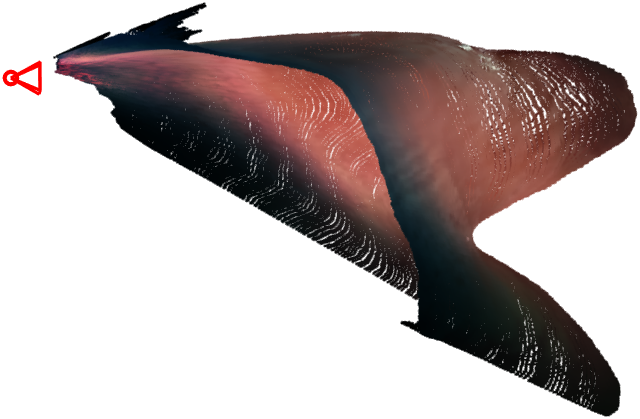}\vspace{-0.07cm}\\
\end{tabular}}
\vspace{-0.3cm}
   \caption{\small{\textbf{Qualitative evaluation on real videos.} Columns show from left to right: arbitrary input frame, RGB rendering, input depth estimation, depth rendering, and side 3D view. From top to bottom: TECAB1, TECAB2, lung lobectomy and bronchoscopy images.}}
\label{fig:final}
\vspace{-0.3cm}
\end{figure}

\textbf{Evaluation and Comparison.} Equipped with the previous depth estimator in combination with the photometric and deformation terms in Eqs.~\eqref{photo_loss}-\eqref{jac_loss}, we define our NeRFscopy baseline. This solution will then be compared to EndoNeRF~\cite{endoNERF}, and our solution after adding the rest of terms until achieving the full formulation (see Eq.~\eqref{final-loss}). Our results are reported in Table~\ref{tab:ablation-losses}. As can be observed, in general, our NeRFscopy model consistently outperforms EndoNeRF~\cite{endoNERF} in several metrics. However, the use of priors does not produce the same effect in all videos, due to the visual information and deformations being very different --in practice, this could be reduced by including prior knowledge, such as the type of intervention to be observed. Note that adding both gradient and smooth terms together improves the NeRFscopy baseline, producing the best solution as possible on average, but when adding the temporal total variation, the results worsen. Since this loss imposes temporal smoothness between frames, the decay in performance could be motivated by artifacts and high-frequency details in the RGB input that make the changes between two frames not smooth, and therefore the proposed regularization is too strong and constraining for the evaluated scenes. A qualitative evaluation of the best NeRFscopy model on average is shown in Fig.~\ref{fig:final}, where the visual quality and 3D reconstruction can be seen. Despite not having a 3D ground truth for comparison, our solution seems physically possible and is compatible with the input image. Furthermore, in Table~\ref{tab:endonerf_results} we provide quantitative evaluation and comparison on the EndoNeRF~\cite{endoNERF} dataset. As can be seen, our approach outperforms the rest in PSNR and LPIPS metrics, becoming very competitive in SSIM, demonstrating its superiority with respect to state-of-the-art methods. Although our method is not real-time (0.14 FPS), we have prioritized efficacy over efficiency and plan to reduce the computational cost in future work.

\begin{table}[t!]
    \centering
    \resizebox{6cm}{!} {
    \begin{tabular}{l|c|c|c|}
         & PSNR$\uparrow$ & SSIM$\uparrow$ & LPIPS$\downarrow$ \\
        \hline
        EndoNeRF~\cite{endoNERF} & 29.831 &	0.925 & \underline{0.081} \\ \hline
        EndoSurf~\cite{endoSURF} & 34.993 & \textbf{0.954} & 0.113 \\ \hline
        LerPlane-32k~\cite{lerplane} & 35.504 & 0.935 &  0.083\\ \hline
        EndoGaussian-monocular~\cite{endogaussian} & \underline{36.429} & \underline{0.951} & 0.089 \\ \hline
        Ours & \textbf{37.204} &	0.949 &	\textbf{0.054} \\ 
        \hline
    \end{tabular}}
    \vspace{-0.2cm}
    \caption{\small{\textbf{Quantitative evaluation on the EndoNeRF dataset}. The table reports PSNR, SSIM and LPIPS for EndoNeRF~\cite{endoNERF}, EndoSurf~\cite{endoSURF}, LerPlane-32k~\cite{lerplane} and EndoGaussian-monocular~\cite{endogaussian}; and our NeRFscopy method. Results shown as {\textbf{best}} and \underline{second best}. } }
    \label{tab:endonerf_results}
    \vspace{-0.3cm}
\end{table}
\renewcommand\fbox{\fcolorbox{red}{white}}

\begin{figure}[t!]
\centering
\resizebox{6.0cm}{!} { 
\begin{tabular}{@{}ccccc@{}}
    \includegraphics[clip, angle=0, height=0.17\linewidth]{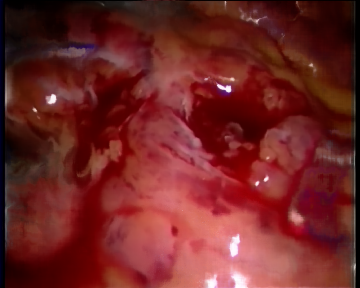}&
     \hspace{-0.6cm}
    {\setlength{\fboxsep}{0pt}
            \setlength{\fboxrule}{2pt}
            \fbox{\includegraphics[clip, angle=0, height=0.17\linewidth]{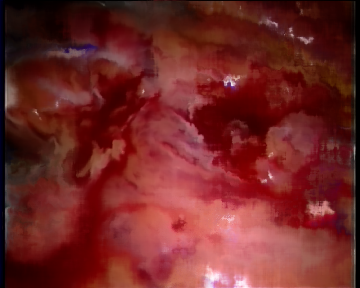}}}&
    \hspace{-0.43cm}
    \includegraphics[clip, angle=0, height=0.17\linewidth]{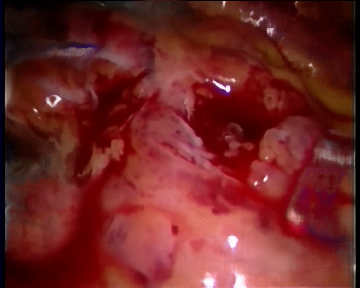}&
    \hspace{-0.6cm}
    {\setlength{\fboxsep}{0pt}
            \setlength{\fboxrule}{2pt}
            \fbox{\includegraphics[clip, angle=0, height=0.17\linewidth]{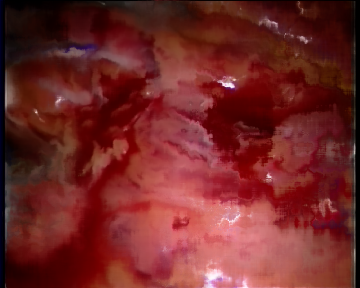}}}&
     \hspace{-0.43cm}
    \includegraphics[clip, angle=0, height=0.17\linewidth]{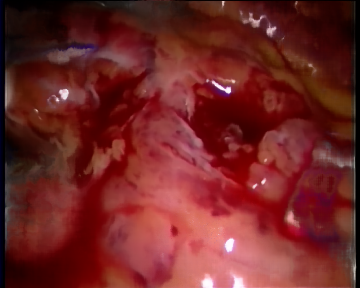}
    \vspace{-0.1cm}\\
    \includegraphics[clip, angle=0, height=0.35\linewidth]{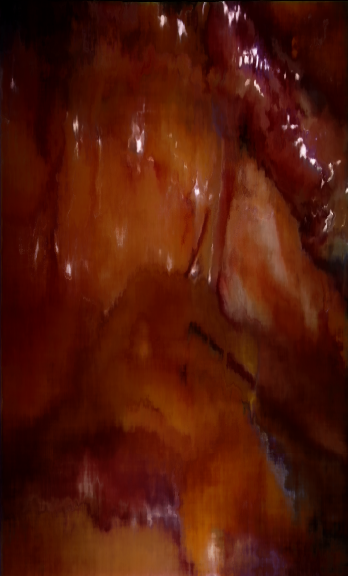}&
     \hspace{-0.6cm}
    {\setlength{\fboxsep}{0pt}
            \setlength{\fboxrule}{2pt}
            \fbox{\includegraphics[clip, angle=0, height=0.35\linewidth]{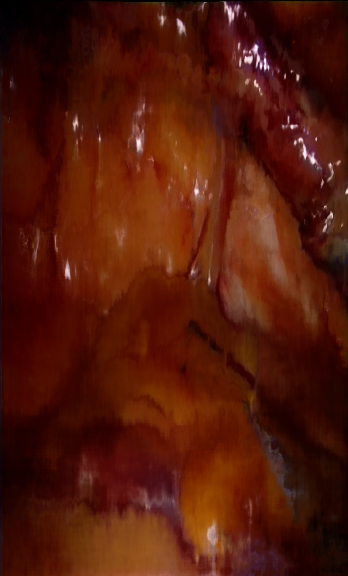}}}&
     \hspace{-0.43cm}
    \includegraphics[clip, angle=0, height=0.35\linewidth]{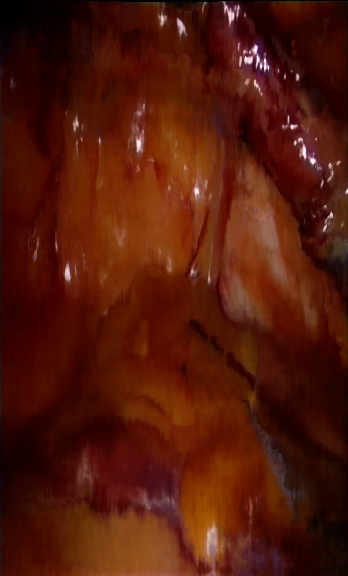}&
     \hspace{-0.6cm}
    {\setlength{\fboxsep}{0pt}
            \setlength{\fboxrule}{2pt}
            \fbox{\includegraphics[clip, angle=0, height=0.35\linewidth]{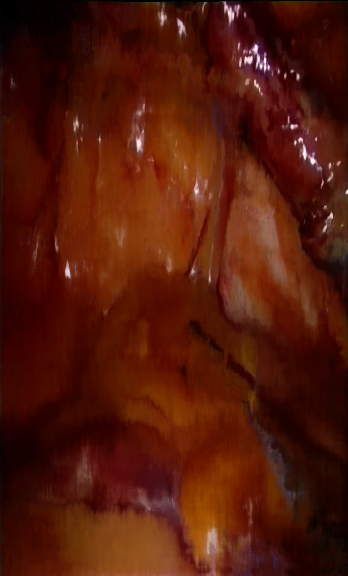}}}&
     \hspace{-0.5cm}
    \includegraphics[clip, angle=0, height=0.35\linewidth]{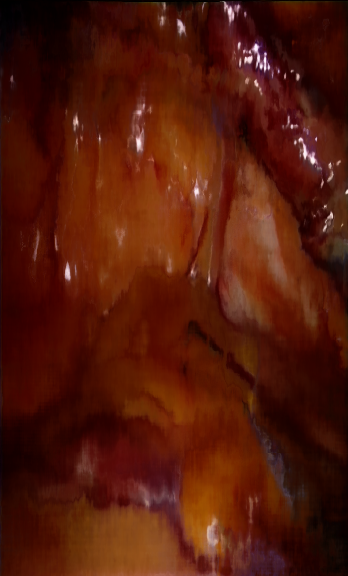}\vspace{-0.1cm}\\
    \includegraphics[clip, angle=0, height=0.35\linewidth]{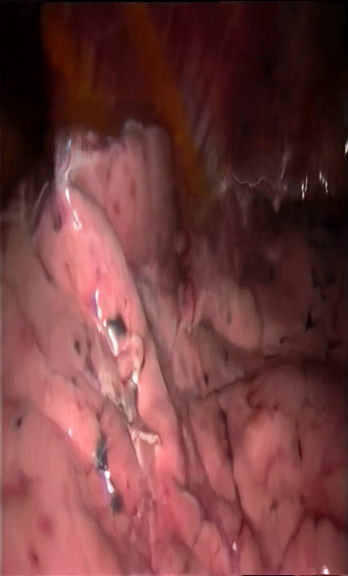}&
     \hspace{-0.6cm}
    {\setlength{\fboxsep}{0pt}
            \setlength{\fboxrule}{2pt}
            \fbox{\includegraphics[clip, angle=0, height=0.35\linewidth]{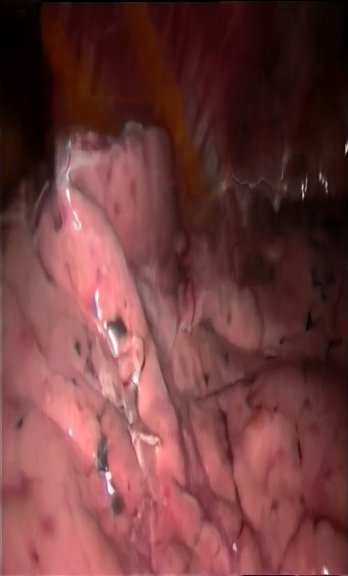}}}&
     \hspace{-0.43cm}
    \includegraphics[clip, angle=0, height=0.35\linewidth]{Figures/novel/sgs/000.rgb.png}&
     \hspace{-0.6cm}
    {\setlength{\fboxsep}{0pt}
            \setlength{\fboxrule}{2pt}
            \fbox{\includegraphics[clip, angle=0, height=0.35\linewidth]{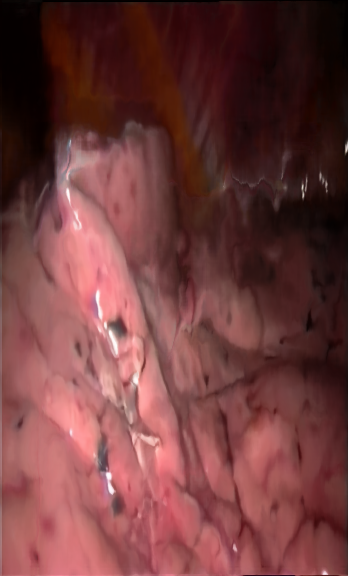}}}&
     \hspace{-0.5cm}
    \includegraphics[clip, angle=0, height=0.35\linewidth]{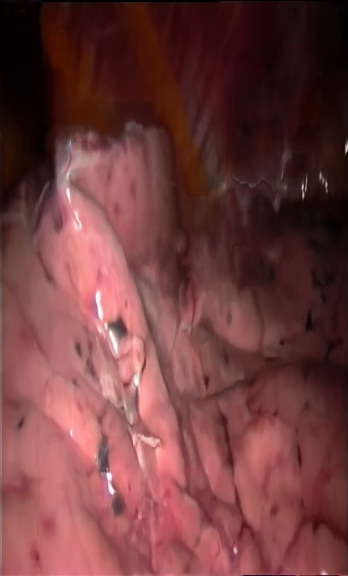}\vspace{-0.1cm}\\

    \includegraphics[clip, angle=0, height=0.17\linewidth]{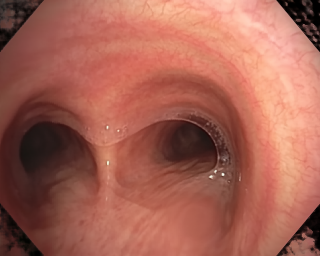}&
     \hspace{-0.6cm}
    {\setlength{\fboxsep}{0pt}
            \setlength{\fboxrule}{2pt}
            \fbox{\includegraphics[clip, angle=0, height=0.17\linewidth]{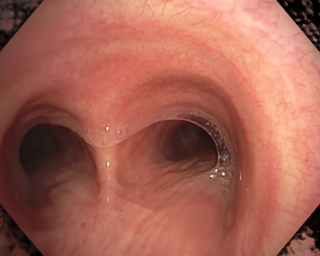}}}&
     \hspace{-0.43cm}
    \includegraphics[clip, angle=0, height=0.17\linewidth]{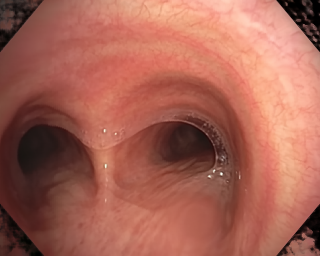}&
     \hspace{-0.6cm}
    {\setlength{\fboxsep}{0pt}
            \setlength{\fboxrule}{2pt}
            \fbox{\includegraphics[clip, angle=0, height=0.17\linewidth]{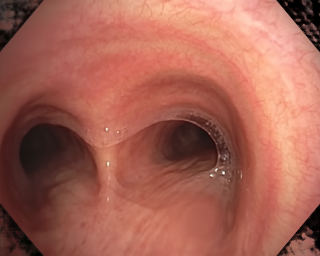}}}&
     \hspace{-0.5cm}
    \includegraphics[clip, angle=0, height=0.17\linewidth]{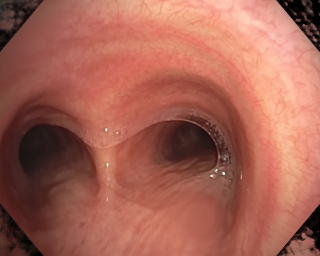}\\
\end{tabular}}
\vspace{-0.3cm}
   \caption{\small{\textbf{Rendered results for five consecutive frames.} Novel views highlighted in red. From top to bottom: TECAB1, TECAB2, lung lobectomy and bronchoscopy images.}}
\label{fig:novelviews}
\vspace{-0.2cm}
\end{figure}

\textbf{Novel view synthesis.} Our NeRFscopy model can be exploited to render novel views (alternative deformation states) that have never been seen in the input video. In Fig.~\ref{fig:novelviews}, we show some instances for the four real videos we use in this study. Again, the visual quality of our rendered estimation is remarkable, while producing physically-aware states.

\section{Conclusion}

We have presented NeRFscopy, a self-supervised and generic method for novel view synthesis of time-varying biological tissues from endoscopy images. To this end, we integrate an SE(3) field to capture tissue deformations in combination with sophisticated terms that allow us to learn an implicit 3D representation of the scene solely from data. We have extensively evaluated our approach on a wide variety of challenging scenarios, obtaining accurate results in comparison with competing approaches. Our future work is oriented to include the camera motion in the formulation.

\bibliographystyle{IEEEbib}
\bibliography{refs}

\end{document}